\documentclass[sigconf]{acmart}

\AtBeginDocument{%
  }

\copyrightyear{2024}
\acmYear{2024}
\setcopyright{acmlicensed}\acmConference[ICMLSC 2024]{2024 The 8th International Conference on Machine Learning and Soft Computing}{January   26--28, 2024}{Singapore, Singapore}
\acmBooktitle{2024 The 8th International Conference on Machine Learning and Soft Computing (ICMLSC 2024), January   26--28, 2024, Singapore, Singapore}
\acmDOI{10.1145/3647750.3647754}
\acmISBN{979-8-4007-1654-6/24/01}

\begin{document}

\title{A Machine learning and Empirical Bayesian Approach for Predictive Buying in B2B E-commerce}

\author{Tuhin Subhra De}
\authornote{Work performed while interning at Udaan.com, a B2B e-commerce company}
\email{tuhinsubhrade@iitkgp.ac.in}
\affiliation{%
    \institution{Indian Institute of Technology}
    \city{Kharagpur}
  \state{West Bengal}
  \country{India}
  \postcode{721302}
}

\author{Pranjal Singh}
\email{pranjal.singh@udaan.com}
\affiliation{%
  \institution{Udaan}
  \city{Bangalore}
  \state{Karnataka}
  \country{India}
  \postcode{560103}
}

\author{Alok Patel}
\email{alokpatel.a@udaan.com}
\affiliation{%
  \institution{Udaan}
  \city{Bangalore}
  \state{Karnataka}
  \country{India}
  \postcode{560103}
}

\renewcommand{\shortauthors}{T.S. De et al.}

\begin{abstract}
In the context of developing nations like India, traditional business-to-business (B2B) commerce heavily relies on the establishment of robust relationships, trust, and credit arrangements between buyers and sellers. Consequently, e-commerce enterprises frequently employ telecallers to cultivate buyer relationships, streamline order placement procedures, and promote special promotions. The accurate anticipation of buyer order placement behavior emerges as a pivotal factor for attaining sustainable growth, heightening competitiveness, and optimizing the efficiency of these telecallers. To address this challenge, we have employed an ensemble approach comprising XGBoost and a modified version of Poisson Gamma model to predict customer order patterns with precision. This paper provides an in-depth exploration of the strategic fusion of machine learning and an empirical Bayesian approach, bolstered by the judicious selection of pertinent features. This innovative approach has yielded a remarkable 3 times increase in customer order rates, showcasing its potential for transformative impact in the e-commerce industry.
\end{abstract}




\maketitle
\section{Introduction}
Established in 2016 with a vision to revolutionize trade in India through technology, Udaan is the country's largest business-to-business (B2B) e-commerce platform. Udaan operates across diverse product categories, including lifestyle, electronics, home \& kitchen, staples, fruits and vegetables, FMCG, pharma, and general merchandise. With a network of over 3 million registered users and 25,000-30,000 vendors and sellers spanning across 900+ cities in India and encompassing more than 12,000 pin codes, Udaan facilitates over 4.5 million transactions per month. In the traditional B2B environment in India, purchasers are situated in both major metropolitan areas and remote rural areas, with a limited level of digital literacy. Transactions in this context are frequently influenced by established relationships, trust, and credit arrangements. To excel in this competitive arena, Udaan has assembled a substantial team of sales executives (telecallers) who are dedicated to nurturing buyer relationships, addressing concerns, helping with order placement, and keeping buyers informed about promotions and sales events.

 In the ever-evolving world of B2B e-commerce, where the delicate interplay of relationships and transactions fuels trade, the ability to predict buyer behavior is a critical asset for sustainable growth and competitiveness. Customer loyalty\cite{Customer-loyalty} and customer retention\cite{churn-prediction} largely hinges on business decisions that simplify their experiences and elevate the quality of service\cite{call-center-customer-satisfaction}. The digital landscape has provided B2B enterprises with a wealth of data and technological tools that have reshaped how business buyers navigate the procurement process. Many organizations employ telecallers to engage customers in product purchases, specializing in reminding customers about items they may have previously encountered on the e-commerce platform. However, due to the complexity of understanding past order history and the factors driving previous purchases, coupled with limited telecaller bandwidth, it is essential to target customers with a high likelihood of making future purchases. This approach not only has a positive impact on revenue but also efficiently utilizes human resources, resulting in significant cost savings.

Previous research has identified significant patterns in customer buying habits. For instance, \cite{MPG} demonstrated that customers tend to repeat their purchasing patterns not only for fast-moving consumables like toothpaste and soaps but also for electronic items such as memory cards and HDMI cables. They proposed various methods to predict the likelihood of customers buying specific items. Some of these methods include Poisson-Gamma and Modified Poisson-Gamma models. Additionally, \cite{RRP_stat} integrated the Poisson-Gamma model with the Dirichlet model to forecast which products customers are likely to purchase. Research in \cite{NBD} has also explored repeat purchasing behaviors, particularly related to product brands. For instance, if we consider a specific brand, like a brand of noodles, researchers have examined how many customers are expected to purchase this brand in fixed time intervals, such as 1 week or 3 weeks. These insights provide valuable information for understanding and predicting customer behavior in the context of product purchases.

Our methodology sets itself apart from prior research by integrating a machine learning model, specifically XGBoost, to forecast potential buyers who are likely to make orders within the upcoming 48 hours. XGBoost is a cutting-edge algorithm renowned for its outstanding performance in various domains, including e-commerce\cite{gan2022xgboost}. Additionally, we incorporate cross-learning capabilities through a modified Poisson-Gamma model, which analyzes the entire history of buyer purchases, product listings, and brand interactions. This combined approach enables us to pinpoint the most probable buyers and generate a ranked list based on their likelihood of placing an order. This fusion of methods significantly enhances the accuracy and coverage of our predictions across the entire customer base of the company. Furthermore, our methodology is adaptable to different timeframes, such as 3 days or 1 week, depending on the specific use case.

In this study, we explore the transformative potential of predictive purchasing models in combination with an empirical Bayesian approach within the realm of business-to-business (B2B) interactions. Our focus is on illuminating how these approaches can contribute to unlocking future sales opportunities and establishing a solid foundation for long-term success. Through the utilization of an ensemble of models trained on extensive historical order data, we have achieved an F1-score of 0.77 on the test dataset, which encompasses customers from various cities across India. Over a 3-week experimental phase, we observed a remarkable increase in the conversion rate of potential buyers, which soared by more than 3x to an impressive 13\%, representing a substantial improvement compared to the previous rate of 4\%. As a result of these promising findings, the conventional system has been replaced by this innovative machine learning-driven system, marking a significant advancement in our approach.

\section{Proposed Approach}
Our research work revolves around the task of gauging the likelihood of customers engaging in product purchases within the next 48 hours. To achieve our goal, we use a special way of looking at time, using repeated 2-day periods to learn from the changes in transactions. The choice of  48-hour time window empowers us to manage both call connection failures as well as attending to the buyer order needs within the right time frame providing a robust foundation for our predictive methodologies.

Our prediction framework is built upon the cutting-edge tree-based model XGBoost, which has won praise for its accuracy in predictive tasks. This model becomes our tool of choice to measure the tendency of a customer to initiate a purchase within the imminent 48-hour window. The careful selection and engineering of input features for XGBoost are guided by a thoughtful hypothesis that attributes a direct influence to these features on a customer's inclination to place an order as well as learnings from ordering patterns in B2B e-commerce. For example, slow moving products have a slower replenishment rates whereas fast moving products have weekly/bi-weekly replenishment rates.

Aiming to further bolster the acuity of our predictions, we adopt an innovative approach characterized by model stacking. Herein, we combine the predictive capabilities of XGBoost with the sophistication of the Modified-Poisson Gamma (MPG) model \cite{MPG}. MPG model helps us to capture the repetitive behaviour of different segment of buyers accurately without much feature engineering. This symbiotic integration endeavors to extract a more comprehensive dynamics underlying customers' interests towards making purchases.

However, the convergence of outputs stemming from these two distinct predictive engines necessitates meticulous handling. To address this, we employ logistic regression as the ultimate classifier or meta-learner, carefully calibrating the weights attributed to the contributions of each individual model. This judicious combination offers a quantified assessment of the likelihood that a given customer will make a purchase within the impending 48-hour time-frame.

\subsection{XGBoost Approach}
\label{ssec:XGBoost-approach}
\begin{figure}[!ht]
    \centering
    \includegraphics[width=\columnwidth]{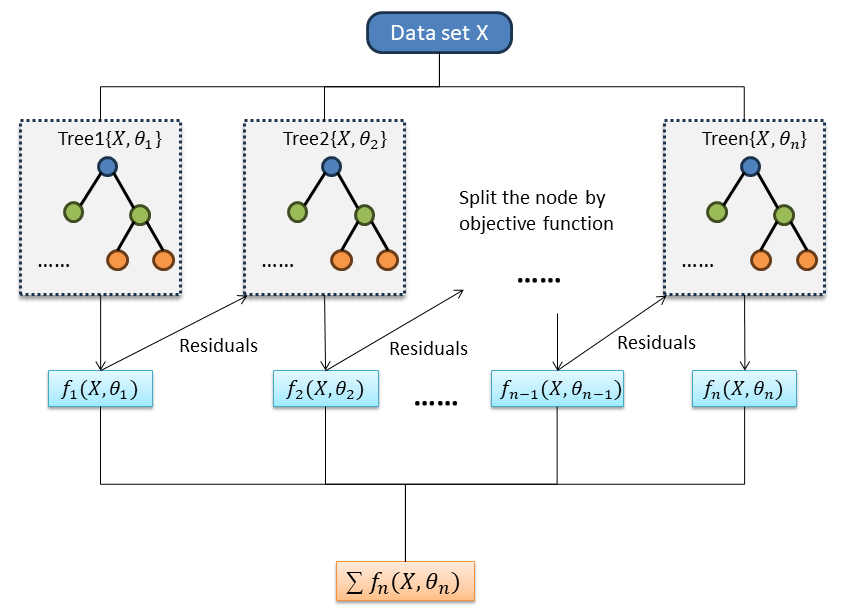}
    \caption{Flow of data in XGBoost model with $n$ trees. The next tree is trained on the residuals of the previous tree, in other way it corrects the mistakes made by the previous one.}
    \label{fig:XGBoost Flow}
\end{figure}

In recent times, tree-based methods have emerged as remarkably effective tools in the realm of predictive analytics, showcasing substantial advancements in both predictive accuracy and adaptability. These non-parametric models, propelled notably by the potent gradient boosting techniques, exhibit a remarkable capacity to readily conform to a diverse array of approximation functions. Among these, XGBoost stands out as a prominent ensemble method that harnesses the collective prowess of multiple weak learners, notably the Classification and Regression Trees (CART). Diverging from the methodology of the Random Forest algorithm, wherein the final output relies on aggregation or majority voting, XGBoost \cite{XGB} follows a distinctive path, employing the gradient boosting approach. In this sequential learning process, each successive learner focuses on mitigating the residuals or errors left behind by its predecessor, thus progressively refining the model's predictive capabilities. This intricate mechanism facilitates the correction of mistakes made by previous learners, resulting in a more robust and accurate predictive model. Figure \ref{fig:XGBoost Flow} shows the flow of data within the learners. The process of adding a new learner continues until the residuals stop reducing or a certain number of iterations are reached. Then a weighted ensemble technique along with the learning rate is used to combine all the learners to give the final output.

XGBoost boasts several advantages over conventional gradient boosting decision trees. Notably, it leverages column sampling, a technique that bolsters its ability to handle high-dimensional data effectively. Moreover, it possesses the capability to transform missing values into a sparse matrix, ensuring the seamless continuation of the tree-building process. These attributes collectively contribute to mitigating the risk of overfitting, a common challenge in predictive modelling. In essence, XGBoost represents a formidable tool in predictive analytics.

Assuming a dataset represented in form of $D = \{(x_i, y_i); i \in \{1, 2, 3 ... n\}; x_i \in \mathbb{R}^w, y_i \in \mathbb{R}\}$. We get $n$ samples with $w$ features each and $y$ being the label. Let $\widehat{y_i}$ be the prediction on sample $i$ then it is given by:

\begin{equation}
    \widehat{y_i} = \sum_{k=1}^K f_k(x_i)
\end{equation}
where $f_k$ is a CART, $K$ is the total number of trees involved in prediction. Since it is iterative process, hence the objective function at a particular iteration say $t$ is given by:
\begin{equation}
    \text{minimize} \sum_{i=0}^{n} L\left(y_i, \widehat{y_i}^{(t-1)} + f_t(x_i)\right) + \omega (f_t)
\end{equation}
where $L$ is an appropriate loss function, $\omega (f_t)$ is the canonical form of tree and $\widehat{y_i}^{(t-1)}$ are the predictions from previous trees. The inputs to this model is the processed feature data and labels, which denotes whether a customer has placed at least an order within the desired time frame or not. A detailed description of data is given in Subsection \ref{ssec:feature-dataset}.

\subsection{Poisson Gamma Model}
In the previous section, we described XGBoost approach to predict likelihood of a buyer placing an order that focused on predicting aggregate customer behavior. While the previous model did not focus on modeling repeat purchase behavior of different buyer segments, the Poisson Gamma model has the capability to generate repeat purchase recommendations for individual customers.

Modeling repeat purchase recommendations involves identifying customer repeat purchases holistically. An efficient approach will be consideration of the number of times a customer repeat purchased a product and their repeat purchase periodicity. Lots of work has been done in formulating the repeat purchase behaviour with Negative Binomial Distribution (NBD) \cite{NBD1}, \cite{NBD2}. Most of them are based on aggregate customer buying behaviour on a long term basis. We focus on customer and product specific patterns. In context to this, NBD model used in \cite{NBD} assumed that a customer’s repeat purchases follow a homogeneous Poisson’s process with repeat purchase rate $\lambda$, meaning successive repeat purchases are independent of each other. The second assumption was a gamma prior on $\lambda$, i.e., assume that $\lambda$ across all customers follows a Gamma distribution with shape $\alpha$ and rate $\beta$. As a result, the NBD model is a Bayesian model where the prior on is a gamma prior and the evidence is distributed as a Poisson distribution. Hence this is stated as the Poisson-Gamma model (PG).

A Bayesian estimate of the customer’s repeat purchase rate is obtained by combining the
prior distribution with customer’s own past purchase history using:
\begin{equation}
\label{PG-model}
    \lambda_{A, C}(t) = \frac{k + \alpha_{A}}{t + \beta_{A}},  t > 0
\end{equation}
where $\alpha_{A}$ and $\beta_{A}$ are the shape and rate parameters of the gamma prior of product $A$; $k$ is the number of purchases of product $A$ by customer $C$; and $t$ is the time elapsed between the first purchase of product $A$ by customer $C$ and the current time.

However, mathematically when we formulate a buying activity; say a customer $C$ bought a product $A$, $k^{th}$ time after passing of $t$ time units since its first purchase. The purchase rate just before and after purchase is given by:

\begin{equation}
\label{eqn-purch_rate-t}
    \lambda_{A, C}(t) = 
    \begin{cases}
        \frac{k - 1 + \alpha_{A}}{(t - \epsilon) + \beta_{A}}, 
        & \text{Before purchase} \vspace{1.0em}\\
        \frac{k + \alpha_A}{(t + \epsilon) + \beta_A}, & \text{After purchase}
    \end{cases}
\end{equation}

Where $\epsilon$ is an infinitesimal small time interval in the neighbourhood of the purchase time $t$. As we focus on the exact time of purchase ($\epsilon \to 0$), we observe that the purchase rate immediately after purchase is greater than before purchase, which is contrasting in reality. Introduction of a modified version of PG model solves this problem.

\subsection{Modified Poisson Gamma model (MPG)}
The initial assumption within the framework of the Poisson-Gamma model may appear counter-intuitive as we saw in Equation \ref{eqn-purch_rate-t}. This occurs because, theoretically, the probability of events occurring in a homogeneous Poisson's process is a constant and time-independent. A Poisson's process is a limiting instance of Bernoulli's processes that has no memory, an extremely low chance of occurrence, and many trails. In practical life the demand of a product for a customer immediately after purchase is low and gradually may increase with time especially with the case of consumables. It is quite probable that the recurrence of purchase behavior is influenced by the time elapsed since the last acquisition, and this behavior is not entirely time-independent. Consequently, a modification to the traditional Poisson-Gamma model was introduced to accommodate these new insights, resulting in what is referred to as the Modified Poisson-Gamma (MPG) process \cite{MPG}. This adapted framework introduces the following key assumptions:
\begin{itemize}
    \item Assume that a customer’s successive purchases are correlated to each other, this process uses a single repeat purchase rate parameter $\lambda$ and assumes that $\lambda$ is dependent on the last time the customer repeat purchased that product. 
    \item This is same as followed in PG process, assuming that $\lambda$ across all customers follows a Gamma distribution with shape $\alpha$ and rate $\beta$.
\end{itemize}
By fitting the maximum likelihood estimates of the purchase rates of repeat customers, the parameters of the product-specific gamma distributions are empirically calculated. Post estimating the parameters the parameter $\lambda$ is calculated by:
\begin{equation}
\label{MPG-lambda}
    \lambda_{A, C}(t) =
    \begin{cases}
         \frac{k + \alpha_A}{t_{purch} + 2 * |t_{mean}-t| + \beta_A}, & 0 < t < 2*t_{mean} \vspace{1.0em}\\
        \frac{k + \alpha_A}{t + \beta_A}, & t\geq2*t_{mean}
    \end{cases}
\end{equation}
Here $t_{mean}$ is the estimated mean repeat purchase time interval between successive purchases of product $A$. The concept of $t_{mean}$ is intuitive with first assumption of MPG. The chance of making a purchase at mean time is high. The purchase rate $\lambda_{A, C}$ increases gradually from $t = 0$ to $t = t_{mean}$, Equation \ref{MPG-lambda} achieves its peak value at $t = t_{mean}$, and decreases gradually from $t = t_{mean}$ to $t = 2*t_{mean}$ and beyond.

Once $\lambda$ is determined, the probability mass function of Poisson distribution for a customer buying the product $m$ times in time interval $t$ is given by :

\begin{equation}
    \label{pmf-poisson}
    P_{A, C}(m) = \frac{\lambda_{A, C}^m\exp(\lambda_{A, C})}{m!}
\end{equation}

To enhance readability, we simplify the notation by omitting the explicit mention of the variable $t$ when referring to $\lambda$. Keep in mind that $\lambda$ is inherently a function of time, and this characteristic extends to Equation \ref{pmf-poisson} as well.

\subsection{XGBoost + MPG approach}
We employ stacking to harness the predictive capabilities of both XGBoost and MPG models. Stacking involves combining the outputs of two or more distinct models or heterogeneous models, culminating in a final learner, often termed the meta learner. These diverse models generate outputs that serve as inputs to the meta learner, ultimately yielding the desired final output.

\begin{figure}[!ht]
    \centering
    \includegraphics[width=\columnwidth]{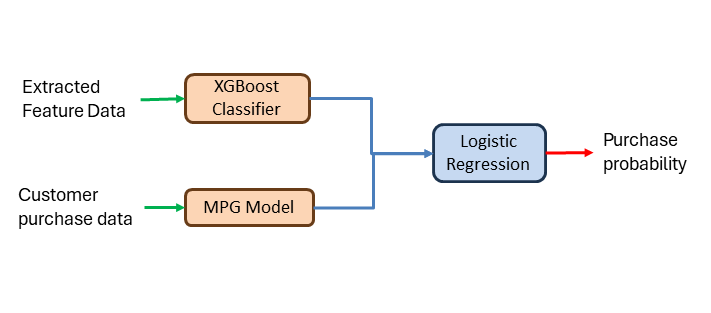}
    \caption{Stacking XGBoost and MPG model with Logistic regression as meta learner}
    \label{stacking fig}
\end{figure}

In our specific scenario, the heterogeneous models consist of MPG and XGBoost, while Logistic Regression assumes the role of the meta learner. Figure \ref{stacking fig} provides a visual representation of the data flow within these models. To provide a bit more detail, XGBoost receives its input as described in Subsection \ref{ssec:XGBoost-approach}, whereas the MPG model takes as input the timely customer past order data encompassing various products. The output of the XGBoost model, which signifies the probability of customer purchases, and the features extracted from the MPG's output are combined as inputs to the Logistic Regression model.

Let us make it clear that the output generated by MPG serves as a foundation for creating additional features with respect to the probability of a buyer placing an order. MPG provides us with probabilities that are specific to each customer and product. Given that our primary goal revolves around identifying whether a customer places at least one order ($m>=1$), we determine this probability by subtracting the probability of no order (when $m=0$) from 1. Therefore, the essential probability, represented by the raw output from MPG, can be expressed as follows:

\begin{equation}
    \label{eqn:prob_atleast_1order}
    P_{A, C}(\text{m>=1}) = 1 - P_{A,C}(m=0)
\end{equation}

Since we are not predicting at product specific level, we aggregate the probabilities by customers across the all the products they interacted with and carve out buyer level features denoting probability of order placement. Table \ref{table-MPG-feature_stats} presents the statistical information of these derived features. To keep most of the information, the new features created at the customer level are:
\begin{itemize}
    \item Number of products with purchase probabilities greater than thresholds $\in$ \{0.5, 0.75, 0.9\}.
    \item Mean, sum and standard deviation of the buyer order probabilities. We use sum and mean simultaneously to have an idea of the number of products a customer was associated with in their entire history.
\end{itemize}

The final purchase probability of a customer within next 48 hours can be termed by:

\begin{equation}
    \label{meta-output}
    P_C = \frac{1}{1+\exp(-(W^T\textbf{x} + b))}
\end{equation}
where $\textbf{x}$ is the concatenated vector of the output from XGBoost regressor and the features from MPG as described in Table \ref{table:feature-info-stats} and Table \ref{table-MPG-feature_stats}. $\textit{W}$ is the weight trainable weight vector and $\textit{b}$ is also trainable. It is the offset term which signifies the purchase probability of customer when $\textbf{x}=\textbf{0}$. The labels used in training of stacked model are the same that were used in standalone XGBoost model (subsection \ref{ssec:XGBoost-approach}).
\section{Results}
In the process of training all the models employed in this work, we employ the compute infrastructure provided by the Microsoft Azure framework within a Python environment. To expedite this computationally intensive work, we leverage a robust compute cluster composed of high-performance Azure \textit{D8asv4} machines, each powered with 32 GB of RAM, $3^{rd}$ Generation AMD EPYCTM 7763v processors (up to 3.5GHz) and uses premium SSD, thus ensuring that our model training proceeds efficiently.

\subsection{Feature Dataset}
\label{ssec:feature-dataset}
\begin{table*}[!ht]
    \centering
    \resizebox{\textwidth}{!}{%
    \begin{tabular}{llllll}
    \hline
        \textbf{Feature} & \textbf{Mean} & \textbf{Std} & \textbf{Min} & \textbf{Median} & \textbf{Max} \\ \hline
        l1wo \textit{(l$\textbf{n}$wo signifies \#Orders placed in the last $\textbf{n}$ week(s))} & 1.04 & 2.39 & 0 & 0 & 103 \\ 
        l2wo & 1.88 & 3.7 & 0 & 0 & 151 \\ 
        l3wo & 2.65 & 4.93 & 0 & 1 & 174 \\ 
        l4wo & 3.38 & 6.14 & 0 & 1 & 206 \\ 
        l5wo & 4.11 & 7.35 & 0 & 1 & 224 \\ 
        l6wo & 4.82 & 8.54 & 0 & 2 & 244 \\ 
        l7wo & 5.53 & 9.71 & 0 & 2 & 268 \\ 
        l8wo & 6.24 & 10.89 & 0 & 2 & 327 \\ 
        DSLO \textit{(\#Days passed since the last order was placed)} & 69.66 & 78.71 & 0 & 20 & 182 \\ 
        l1w\_fam \textit{(l$\textbf{n}$w\_fam signifies \#FAM* visits to customer stores in last $\textbf{n}$ week(s))} & 0.44 & 1.15 & 0 & 0 & 84 \\ 
        l2w\_fam & 0.77 & 1.8 & 0 & 0 & 130 \\ 
        l3w\_fam & 1.1 & 2.41 & 0 & 0 & 132 \\ 
        l4w\_fam & 1.41 & 3.01 & 0 & 0 & 133 \\
        call\_engage\_frac \textit{(Fraction of time engaged with human customer service agent)} & 0.54 & 0.45 & 0 & 0.87 & 1 \\ 
        bot\_engage\_frac \textit{(Fraction of time engaged with bot customer service agent)} & 0 & 0.01 & 0 & 0 & 1 \\ 
        credit\_ratio \textit{(Ratio of credit limit left to total limit allotted)} & 0.05 & 0.32 & -30.04 & 0 & 2.63 \\ 
        promise\_diff \textit{(\#Days delivery date deviated from actual promised date)} & 4.69 & 10.03 & -3 & 1 & 15 \\ 
        max\_promise\_diff & 6.52 & 9.41 & -3 & 4 & 16 \\ 
        min\_promise\_diff & 3.07 & 11.89 & -3 & 0 & 15 \\ 
        a2c\_amount \textit{(Worth of all items in add to cart till date)}& 141 & 32020.15 & 0 & 0 & 5142694 \\ 
        l7d\_a2c \textit{(Worth of highest frequently item in add to cart since last 1 week)} & 106.3 & 31961.6 & 0 & 0 & 5142694 \\ 
        l7d\_a2c\_2 \textit{(Worth of 2nd highest frequently bought item added to cart in last 1 week)} & 63.08 & 31279.51 & 0 & 0 & 5142694 \\ 
        num\_app\_opens\_l7d \textit{(\#app was opened in last 1 week)}& 11.45 & 15.41 & 0 & 7 & 2223 \\ 
        num\_listing\_views\_l7d \textit{(\#List views in last 1 week)}& 4.83 & 7.75 & 0 & 3 & 1096 \\ 
        num\_a2c\_l7d \textit{(\#items added to cart in last 1 week)} & 1.42 & 2.42 & 0 & 1 & 91 \\ \hline
    \end{tabular}%
    }
    \caption{Descriptive statistics of the numerical features used for XGBoost algorithm. *FAM stands for Field Account Manager, the personnel who visits customer stores for solving queries and maintenance of good relation. Min and Max stands for the minimum and maximum values respectively, Std is the standard deviation.}
    \label{table:feature-info-stats}
\end{table*}

The entirety of the data utilized in this research work is the exclusive property of \textit{Udaan.com} and has been meticulously sourced from their comprehensive data warehouse. Within this dataset, the feature data is temporally lagged by a period of 2 days with respect to the labels, which are binary in nature. Specifically, a label of '1' denotes whether a customer has placed at least one product order within the subsequent 2 days, while a '0' signifies otherwise. Additionally we provide descriptive statistics of the various features in our dataset in Table \ref{table:feature-info-stats}


It's imperative to note that the initial real-time dataset that was extracted exhibited a substantial class imbalance. In response, we employed an undersampling technique to rectify this imbalance, involving the random removal of data points from the majority class. The transformation is detailed in Table \ref{sample-table}, which provides insights into the characteristics of the labels both before and after the undersampling process. Other procedures include Tomeklinks \cite{Ivan-CNN}, which primarily focuses on removal of majority class samples that are in vicinity of the minority class increasing the segregation of the classes. This method, numerically leads to very low down-sampling of majority class in number which is contrasting to our scenario. The ratio of majority to minority in raw data is about 80:1 which demands for high scale undersampling. An interesting methodology proposed by \cite{wilson-NN} used neighbour comparison. The neighbours which are in close proximity and do not comply with the others in prediction are removed. This preserves important information, but is highly computationally expensive. Given the size our majority sample set and mentioned factors, we decided to follow random undersampling technique as the best fit technique in this case.  Notably, the abundance of samples post-undersampling ensures that we do not compromise the integrity of the dataset while mitigating the class imbalance.

\begin{table}[!ht]
    \centering
    \resizebox{0.9\columnwidth}{!}{
    \begin{tabular}{c|cc}
    \hline
        \textbf{Samples} & \textbf{Before sampling} & \textbf{After sampling}  \\ \hline
        \# 0's & 2,767,563 & 69,144 \\ 
        \# 1's & 34,572 & 34,572 \\ 
        Total & 2,802,135 & 103,716 \\ \hline
    \end{tabular}
    }
    \caption{Number of samples belonging to each class in the whole dataset before and after sampling.}
    \label{sample-table}
\end{table}

\begin{figure}[!ht]
    \centering
    \includegraphics[width=\columnwidth]{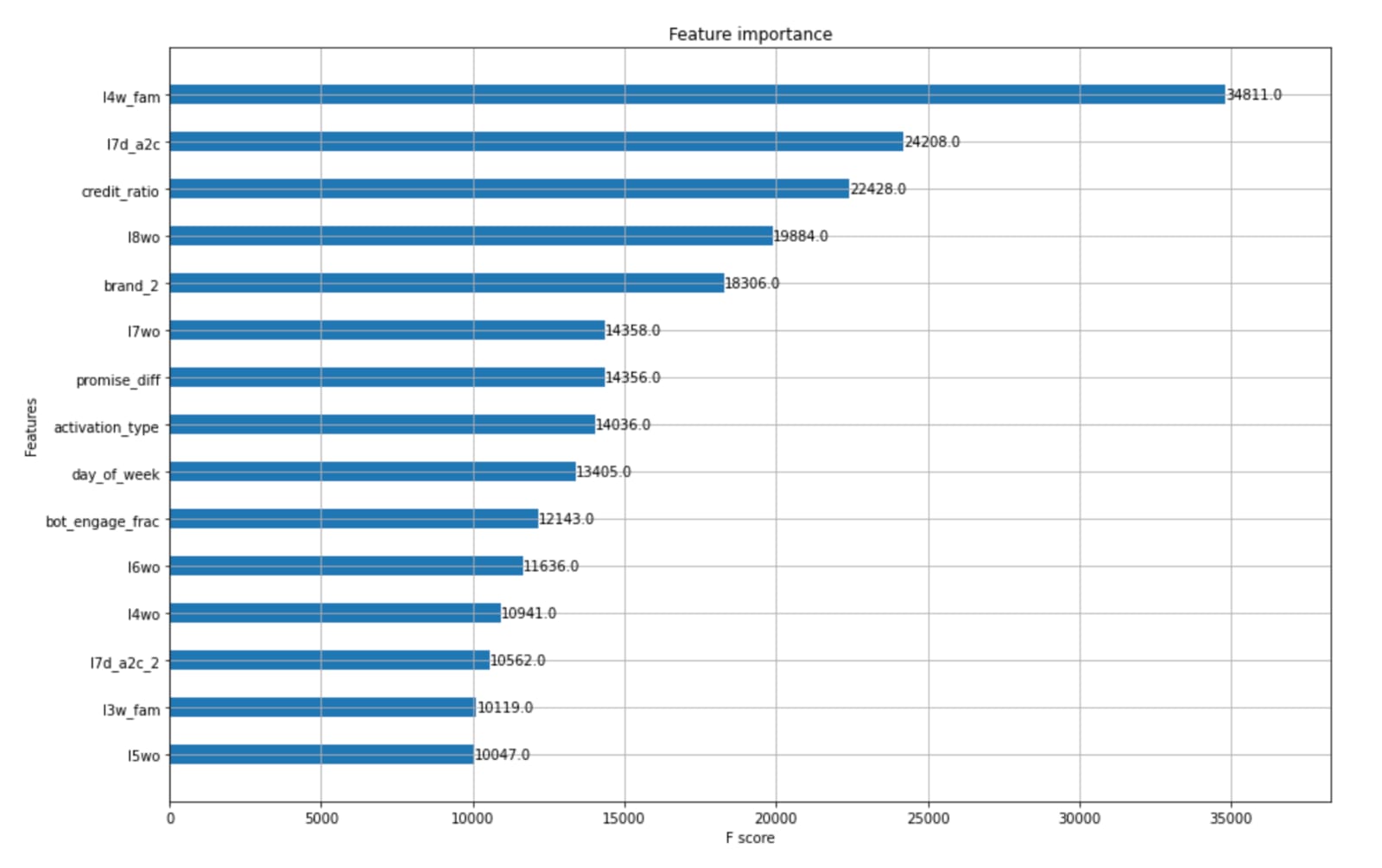}
    \caption{Top 15 input features in decreasing order of importance obtained from standalone XGBoost prediction.}
    \label{fig:fimp}
\end{figure}

In addition to the numerous categorical features, our dataset encompasses variables such as the type of credit \textit{(activation\_type)}, the specific day of the week when customers placed the maximum number of orders \textit{(day\_of\_week)}, and the names of the most frequently purchased brands, including the highest \textit{(brand\_1)}, second-highest \textit{(brand\_2)}, and third-highest \textit{(brand\_3)}. These categorical attributes were instrumental in our analysis.

The selection process focused on identifying the top 15 features, which are showcased in Figure \ref{fig:fimp}, arranged in descending order of their significance based on their decrease in impurity when splitted. Notably, we observe many numerical features being the most informative ones. An interesting feature \textit{bot\_engage\_frac}, depicting the ratio of amount of time a customer being engaged with bot to the total time they spent on customer service, makes in top 15. This feature might be negatively correlated with our objective but shows that quality of customer service can influence businesses to a lot extent \cite{call-center-customer-satisfaction}.

Subsequently, we partitioned the dataset into a train-validation-test split with an 80:5:15 ratio, signifying the allocation of data for training, validation and testing respectively. All training tasks, including the fine-tuning of hyperparameters, were executed exclusively on the training and validation subset respectively. The best-fitting model, as determined during this process, was then employed to make predictions on the test set, ensuring an unbiased evaluation of model performance.

\subsection{Standalone XGBoost performance}
\label{sec:xgb-solo-results}
The XGBoost model is driven by two fundamental components: the meticulously prepared feature data and the corresponding labels. In our quest to pinpoint the hyperparameters that best align with our dataset and enhance model performance, we employ a thorough 5-fold grid search cross-validation strategy. This comprehensive approach involves systematic exploration and tuning of hyperparameters, encompassing considerations such as the number of trees, the learning rate, the maximum tree depth, and the maximum number of leaves within each tree. To delve deeper into harnessing the impact of features across various levels, we have incorporated a multifaceted approach. This includes column sampling during the tree construction process, at various tree depths, and even during node splits.

\begin{figure}[!ht]
    \centering
    \includegraphics[width=\columnwidth]{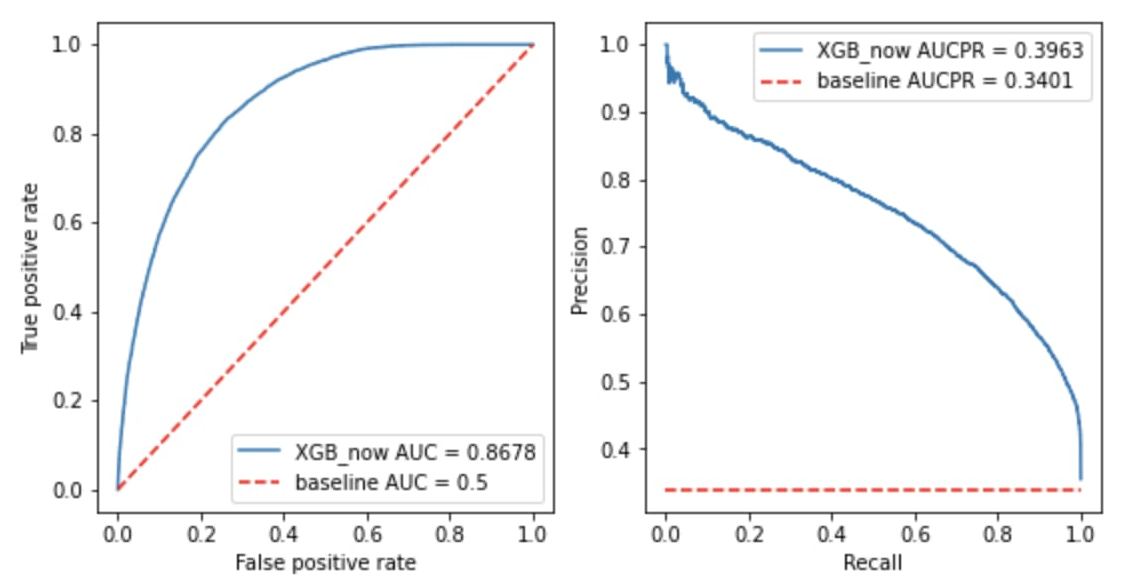}
    \caption{The AUC-ROC and AUC-PR of XGBoost model}
    \label{fig:XGB-AUCPR}
\end{figure}

Given the nature of our classification task, we have fine-tuned the parameter \texttt{eval\_metric} to focus on the area under the precision-recall curve, ensuring that our model's performance aligns with our business objectives. It is important to note that our training schedule is designed to adapt to the dynamic nature of customer data. While customer information undergoes daily updates, our model undergoes training at a frequency of once every three weeks. However, the predictive model is actively employed every two days to provide real-time insights.

Leveraging the computational capabilities of the earlier specified cluster, the entire training process is efficiently completed within a total duration of 3 hours. This streamlined approach enables us to stay agile and responsive to changing customer behaviors while ensuring that our predictive model remains accurate and effective.

Theoretical results of XGBoost's performance is presented in Figure \ref{fig:XGB-AUCPR}. We obtain an AUC-ROC score of 0.8678 and AUC-PR score of 0.3963.

\subsection{Stack model performance}
The stack model undergoes its own training regimen, where we fine-tune its hyperparameters through a 5-fold grid search cross-validation process. Notably, during this phase, we focus exclusively on adjusting the hyperparameters associated with XGBoost and Logistic Regression, while the MPG model remains unchanged. It's worth noting that the MPG model has been trained independently prior to this phase and is kept static during the stack model's hyperparameter tuning.

\begin{table}[!ht]
    \centering
    \resizebox{0.9\columnwidth}{!}{
    \begin{tabular}{lllll}
    \hline
        \textbf{Feature} & \textbf{Mean} & \textbf{Std} & \textbf{Min} &  \textbf{Max} \\ \hline
        \#p\_greater than 0.5 & 0.01 & 0.18 & 0 &  55 \\ 
        \#p\_greater than 0.75 & 0 & 0.06 & 0 &  12 \\ 
        \#p\_greater than 0.9 & 0 & 0.02 & 0 &  7 \\ 
        p\_mean & 0.01 & 0.03 & 0 &  1 \\ 
        p\_std & 0.04 & 0.06 & 0 &  1.53 \\ 
        p\_sum & 0.01 & 0.06 & 0 &  5.26 \\ \hline
    \end{tabular}
    }
    \caption{Descriptive statistics of features obtained from MPG by aggregating the purchase probabilities across all the products for each customer. We add the prefix \textit{p} to the features conveying that the metric is obtained from probabilities in order to avoid confusion with their statistical measures.}
    \label{table-MPG-feature_stats}
\end{table}

\begin{figure}[!ht]
    \centering
    \includegraphics[width=\columnwidth]{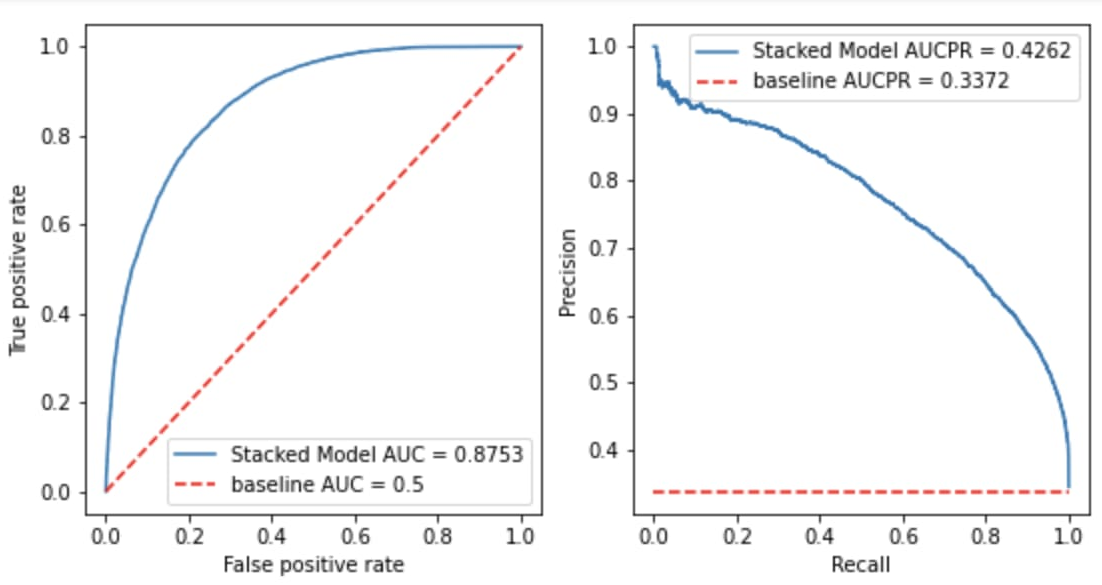}
    \caption{The AUC-ROC and AUC-PR of Stacked model}
    \label{fig:stacked-AUCPR}
\end{figure}

To avoid over-fitting we experiment with regularized version of Logistic Regression. We use elastic penalty and leverage the strength as a hyper-parameter. Since our sample count was much larger than the feature count,  Newton-Cholesky solver is used for quadratic and finer convergence. The entire process took approximately 4 hours to complete on the compute cluster. Figure \ref{fig:stacked-AUCPR} shows the associated curves with the stacked model. We observe a 0.8753 AUC-ROC score along with a 0.4262 AUC-PR score. 

\begin{table*}[!ht]
    \centering
    \begin{tabular}{ccccccc}
    \hline
        \textbf{Models} & \textbf{Accuracy} & \textbf{F1 score} & \textbf{Precision} & \textbf{Recall} & \textbf{AUC} & \textbf{AUC-PR} \\ \hline
        XGBoost & 0.79 & 0.76 & 0.77 & 0.75 & 0.8678 & 0.3963 \\ 
        Stacked & 0.80 & 0.77 & 0.78 & 0.76 & 0.8753 & 0.4262 \\ \hline
    \end{tabular}
    \caption{Theoretical comparative study between the XGBoost approach and Stacked approach.}
    \label{Comparison Table}
\end{table*}

\begin{figure}[!ht]
    \centering
    \includegraphics[width=\columnwidth]{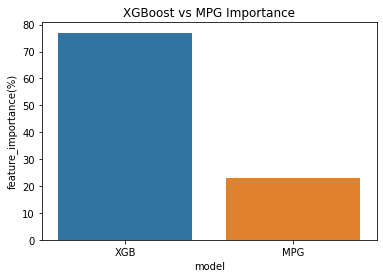}
    \caption{Comparison of feature importance of XGBoost and MPG model in the final stacked model}
    \label{fig:stacked-importance}
\end{figure}

To assess the importance of features coming from XGBoost and MPG we compare their contribution using softmax normalized weights of the meta learner. Figure \ref{fig:stacked-importance} shows their relative contribution in percentage. The feature importance in case of MPG denotes ratio of the sum of normalized MPG features' weights to all the weights of Logistic regression model.  

\subsection{Comparative studies}
\label{ssec:comparative studies}
\subsubsection{Theoretical comparison}

We present a comprehensive analysis comparing the two approaches in Table \ref{Comparison Table}. Our observations indicate that the stacked model surpasses the standalone XGBoost model in terms of performance. Specifically, we note a notable improvement in the AUC-PR score, with a relative increase of 7.5\%, as well as a 1.3\% enhancement in the F-1 score. While these numerical improvements might seem modest, we will later see that in our real-world business application, the stacked model's superior performance becomes more evident and consequential.

To get a concrete idea of classification, we select a threshold of 0.5 on purchase probability. The samples with probability greater than 0.5 are assigned 1 else 0. We denote True Positives, True Negative, False positives and False Negatives by \(TP\), \(TN\), \(FP\), \(FN\) respectively. The general metrics involved in classification problems are accuracy, True positive rate (recall), Precision and F-1 Score. The calculations of these are given by:

\begin{equation}
    accuracy = \frac{TP+TN}{TP+TN+FP+FN}
\end{equation}

\begin{equation}
    precision = \frac{TP}{TP+FP}
\end{equation}

\begin{equation}
    recall = \frac{TP}{TP+FN}
\end{equation}

\begin{equation}
    F1-score = \frac{2TP}{2TP+FP+FN}
\end{equation}
\vspace{0.2 em}
\subsubsection{Comparison by applying in real-time business}
Up to this point, we've presented a comprehensive overview of the theoretical results achieved by our models. However, from a business perspective, our primary objective is to gauge the real-world impact of these approaches, particularly in the context of customer retention—a critical concern for sales revenue growth. Our aim is to assess the effectiveness of these models in boosting sales revenue by tailoring personalized offers to individual customers. To achieve this, we transition from theory to practical application by deploying these models into production, where the telecallers generate call prompts targeting a selected subset of customers. Subsequently, we compare the outcomes of these efforts with the results achieved using existing methods.

This experiment spans a duration of three weeks and encompasses a diverse array of customer segments. Before the commencement of the experiment, we undertake the training of a standalone XGBoost model, followed by the MPG model. Subsequently, we proceed to train the stacked model separately, keeping the MPG model unchanged and allowing the hyper-parameters of the new XGBoost and Logistic Regression models to be tuned. It's worth noting that Equation \ref{eqn:prob_atleast_1order} is inherently time-dependent. As a result, we consistently retrain the MPG model every two days to ensure its alignment with the evolving customer data. Conversely, the standalone XGBoost and stacked models undergo training less frequently. This setup accommodates the dynamic nature of our customer database, enabling us to fetch predictions based on recent data every two days.

\begin{table}[!ht]
    \centering
    \resizebox{\columnwidth}{!}{
    \begin{tabular}{ccccccc}
    \hline
       \textbf{ Models} & \multicolumn{1}{p{2cm}}{\centering \textbf{Probability Range (in \%)}} & 
        \multicolumn{1}{p{2cm}}{ \centering \textbf{\#Customers called}} & 
        \multicolumn{1}{p{1.5cm}}{\centering \textbf{\#Orders in 48 hours}} & \textbf{\% Orders} \\ \hline
       ~ & 90-100 & 9 & 1 & 11.11 & ~ & ~ \\ 
        Non-Model & 80-90 & 165 & 4 & 2.42 & ~ & ~ \\ 
        ~ & 70-80 & 256 & 13 & 5.08 & ~ & ~ \\ \hline
        Total & - & 430 & 18 & \textbf{4.19} & ~ & ~ \\ \hline
        ~ & 90-100 & 11 & 3 & 27.27 & ~ & ~ \\ 
        XGBoost & 80-90 & 157 & 10 & 6.37 & ~ & ~ \\ 
        ~ & 70-80 & 291 & 25 & 8.59 & ~ & ~ \\ \hline
        Total & - & 459 & 38 & \textbf{8.28} & ~ & ~ \\ \hline
        ~ & 90-100 & 12 & 5 & 41.67 & ~ & ~ \\ 
        Stacked & 80-90 & 149 & 18 & 12.08 & ~ & ~ \\ 
        ~ & 70-80 & 287 & 37 & 12.89 & ~ & ~ \\ \hline
        Total & - & \textbf{448} & \textbf{60} & \textbf{13.39} & ~ & ~ \\ \hline
    \end{tabular}
    }
    \caption{Telecalling statistics for each model in real business world. The numbers are aggregated over the course of 3 weeks. Stacked model outperforms others with conversion rate of 13.39\%.}
    \label{RAM calling comparision}
\end{table}

The predictive models generate probabilities representing customers' likelihood of making a purchase. Subsequently, we categorize customers into distinct probability ranges. For instance, customers falling within the 0.9-1.0 probability range might be classified as high-potential customers, and so forth. The real-world performance of these models is meticulously evaluated and presented in Table \ref{RAM calling comparision}.

In the context of telecalling, where bandwidth and manpower resources are constrained, we have opted to focus our efforts on customers whose purchase probability exceeds the threshold of 70\%. This strategic selection streamlines our telecalling efforts, ensuring that we prioritize high-potential customers. It's noteworthy that the manual method, which precedes the introduction of this approach, has exhibited an average conversion rate of 4.19\%. The manual method represents a rudimentary and baseline approach employed within the organization. It relies on simplistic heuristics, such as sequencing customers in descending order based on certain feature data, devoid of any AI capability.

Conversely, the conversion rate experiences a substantial boost, reaching 8.28\% with the solo XGBoost approach. The pinnacle of performance is attained with the stacked model, which achieves an impressive 13.4\% conversion rate. This stark improvement underscores the advantages of employing a stacked architecture, enabling us to harness both feature-level and score-level information to enhance the effectiveness of our customer targeting strategies.
\section{Conclusion}
The predictability of customer purchase behavior holds paramount significance in the realm of e-commerce companies. The ability to identify high-potential customers constitutes a pivotal element in facilitating personalized recommendations and informed decision-making within these companies.

In this paper, we conduct a comparative analysis between a pure machine learning approach and a hybrid machine learning cum parametric-empirical Bayesian approach for predicting customer purchase probabilities. Theoretically, our stacked approach exhibits a modest edge over the standalone XGBoost method. However, when subjected to real-world testing, its performance shines, delivering an impressive 50\% relative increase in aggregate conversion rate, from 8\% achieved by the former method to 13\% by the stacked model. Overall, we see a 3X jump in buyer conversion rate as compared to the earlier manual approach and hence this new introduced model has replaced the earlier telecalling system working on manual decision-making.

It is important to note that this work is presently confined to customer-level predictions. Given that the MPG approach inherently accommodates product-level predictions, there exists a promising avenue for extending this work to encompass both customer and product-level specifications. Achieving this extension might entail the integration of improved machine learning algorithms, specifically product recommendation algorithms and other parameter-estimated models in conjunction with the MPG model.

Furthermore, in selected scenarios, the incorporation of real-time, raw, and direct IOT-based data from customers, as highlighted in \cite{IOT-fuzzy}, holds the potential to facilitate the development of instant fuzzy decision-making models rooted in deep learning. Such integration has the capacity to significantly enhance the quality of predictions and the overall serviceability of the system. This avenue warrants exploration as part of future research endeavors.

\bibliographystyle{ACM-Reference-Format}
\bibliography{citations}

\end{document}